\documentclass[10pt,twocolumn,letterpaper]{article}
\pdfoutput=1
\usepackage{iccv}
\usepackage{times}
\usepackage{epsfig}
\usepackage{graphicx}
\usepackage{caption}
\usepackage{amsmath}
\usepackage{amssymb}
\usepackage[linesnumbered,ruled,vlined]{algorithm2e}
\usepackage{algorithmic}
\usepackage{bm}
\usepackage{booktabs}
\usepackage{tabu}
\usepackage{multirow}
\usepackage{float}
\usepackage{svg}

\usepackage[breaklinks=true,bookmarks=false]{hyperref}

\iccvfinalcopy 

\def\iccvPaperID{****} 

\ificcvfinal\fi

\begin{document}

\title{Class-Guided Image-to-Image Diffusion: Cell Painting from Brightfield Images with Class Labels}

\author{Jan Oscar Cross-Zamirski$^{1,2}$ \hspace{1pt} Praveen Anand$^2$ \hspace{1pt} Guy Williams$^2$ \hspace{1pt} Elizabeth Mouchet$^2$ \hspace{1pt}  Yinhai Wang$^2$ \\
 Carola-Bibiane Schönlieb$^1$ \\
\normalsize{$^1$ DAMTP, University of Cambridge, $^2$ Discovery Sciences, R\&D, AstraZeneca} \\
\textcolor{magenta}{\texttt{\normalsize{\url{https://github.com/crosszamirski/guided-I2I}}}}}

\twocolumn[{%
\maketitle
\renewcommand\twocolumn[1][]{#1}%
\begin{center}
    \centering
    \captionsetup{type=figure}
    \includegraphics[width=1\textwidth]{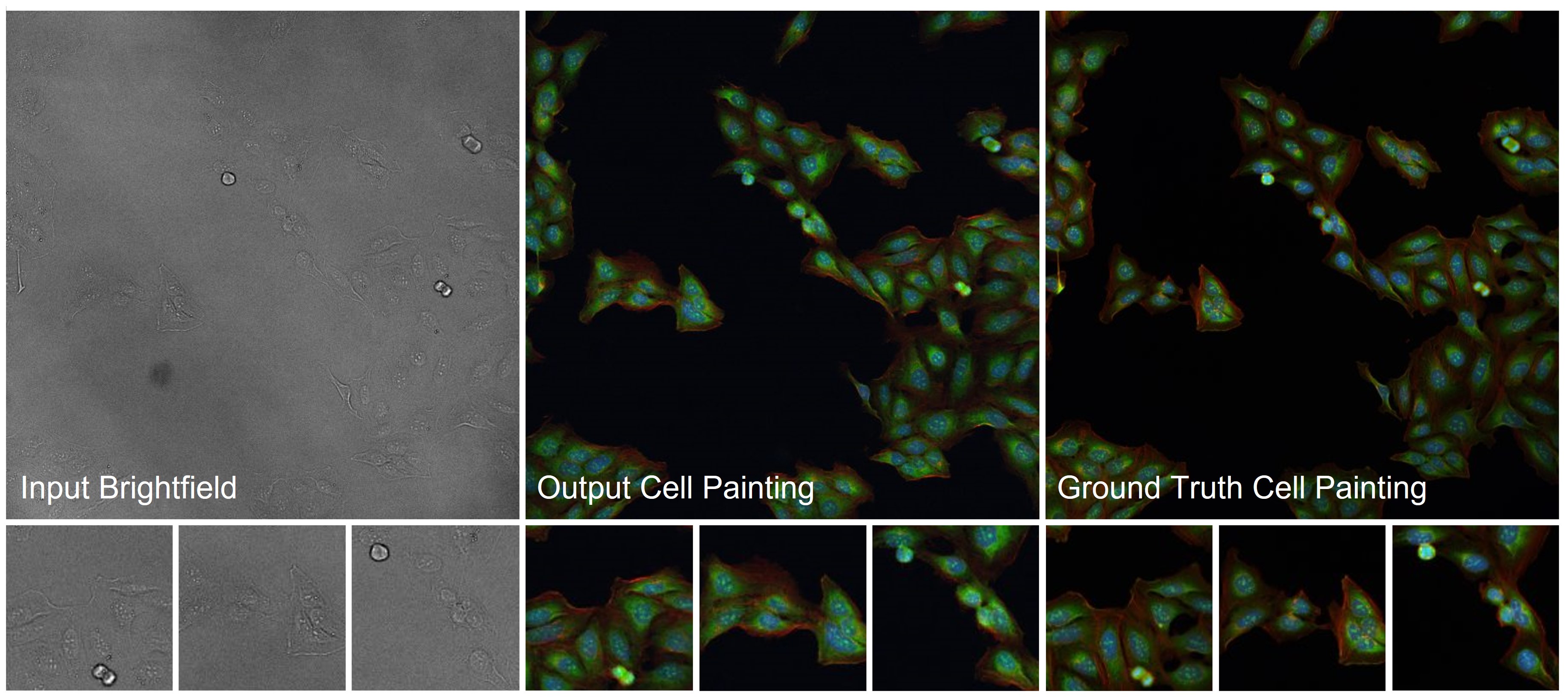}
    \captionof{figure}{A colour composite example of three channels from a test plate: red (AGP), green (ER) and blue (DNA).}
    \label{fig:fig1}
\end{center}%

}]


\begin{abstract} \vspace*{-0.4cm}
Image-to-image reconstruction problems with free or inexpensive metadata in the form of class labels appear often in biological and medical image domains. Existing text-guided or style-transfer image-to-image approaches do not translate to datasets where additional information is provided as discrete classes. We introduce and implement a model which combines image-to-image and class-guided denoising diffusion probabilistic models. We train our model on a real-world dataset of microscopy images used for drug discovery, with and without incorporating metadata labels. By exploring the properties of image-to-image diffusion with relevant labels, we show that class-guided image-to-image diffusion can improve the meaningful content of the reconstructed images and outperform the unguided model in useful downstream tasks.
\end{abstract}

\section{Introduction}

Conditional denoising diffusion probabilistic models (DDPMs) \cite{Ho2020, Dhariwal2021} are trained to learn a probability distribution capable of generating realistic samples from an input condition. These constructions typically fall into one of two categories: models conditional on an input image (image-to-image) \cite{Saharia2021_palette} \textbf{or} models conditional on a class label \cite{Dhariwal2021, Saharia2021}. While many other diffusion models exist which incorporate natural language text encoders such as CLIP \cite{Radford2021} (text-to-image) \cite{Ramesh2022, Saharia2022}, there has been much less attention on advancing models with both paired image \textbf{and} class label information. This can be attributed to a lack of generalist datasets which have both class labels and paired images, as this information can be expensive, sparse or narrow in application \cite{Wu2018, Zhan2022}. 

Despite this, image-to-image problems with discrete metadata appear often in biological and medical image reconstruction. Examples of these inverse problems include PET reconstruction from MRI \cite{Sun2019}, predicting fluorescent labels from transmitted light microscopy \cite{christiansen2018}, sparse-view CT reconstruction and artifact removal \cite{Song2022}. Through the nature of image acquisition there is often additional inexpensive \textit{side} \cite{Sun2019} or \textit{weak label} \cite{caicedo2018} information which can be incorporated to guide the training of the inverse process towards the main task. 
For biological and medical datasets, class labels have been used in deep learning architectures to learn more faithful and generalisable representations \cite{Wang2021, Lin2022}, and as extra information in image-to-image tasks \cite{Sun2019}. 

\begin{figure*}
  \centering
  \includegraphics[width=1\textwidth]{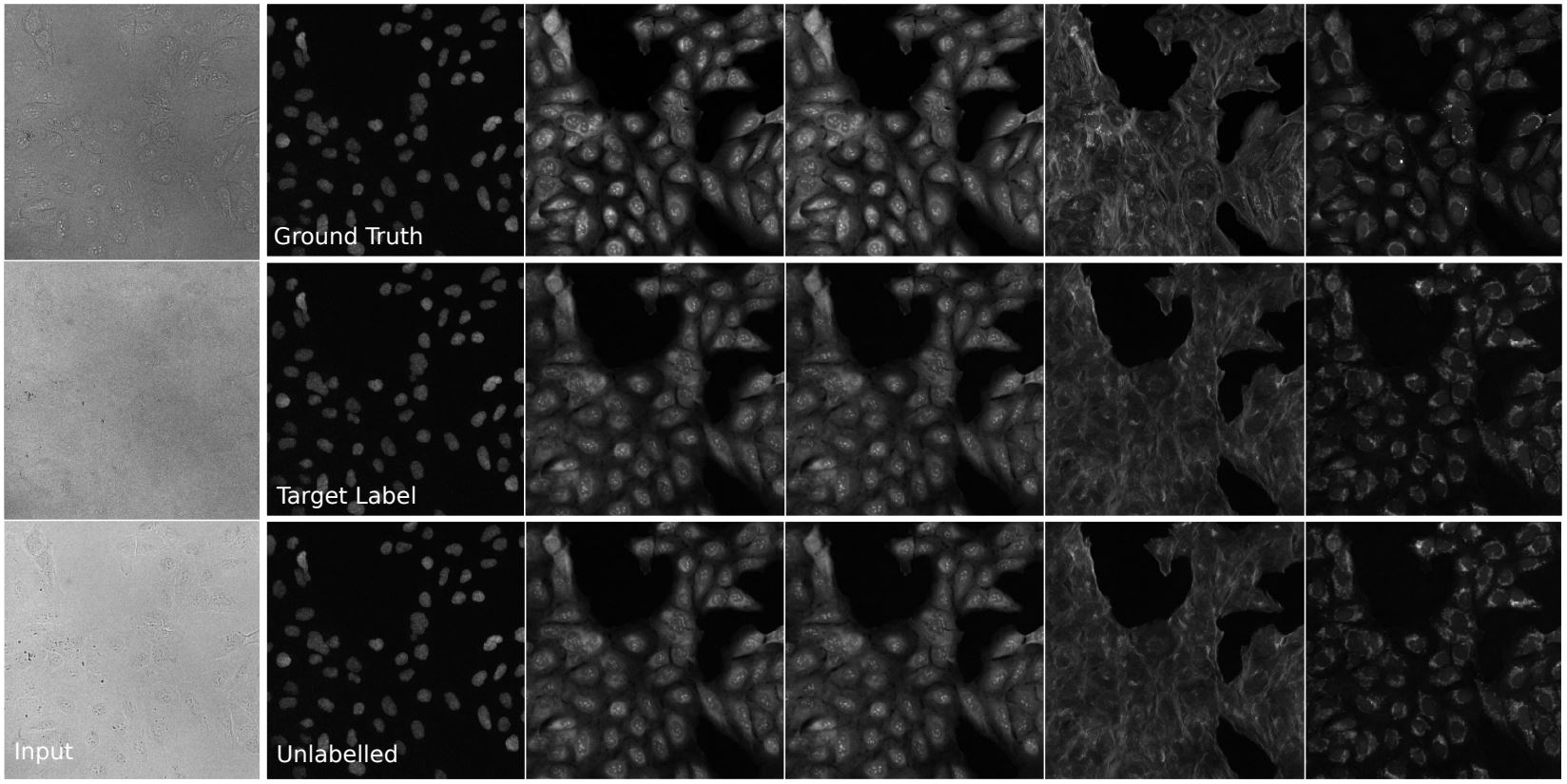}
  \caption[Example of all channels (and input) for both models]{Given input Brightfield (3 channels) our model is able to generate 5 Cell Painting channels. Incorporating meaningful labels can improve biological feature quality and performance on downstream tasks without significantly reducing image quality or adding background noise. Columns left to right: Brightfield (input), DNA, RNA, ER, Mito, AGP.}
  \label{lch61}
\end{figure*}

These problems are dataset specific, and well-established databases of natural images \cite{deng2009, Zhou2017} and their associated labels are rarely analogous to the challenges presented by biological and medical datasets. 
These real-world datasets can have unique types of metadata labels, and application-specific ways to evaluate performance in downstream tasks. Using the predicted images in such tasks to quantify performance may be more important and informative than benchmarking with metrics such as Fréchet Inception Distance (FID) \cite{Heusel2017} and structural similarity index (SSIM) \cite{hore2010}. Accurately capturing a distribution of images is complementary, if not subsidiary, to being able to differentiate between images and their features \cite{Chandrasekaran2021}.

We investigate the utility of image-to-image diffusion with class labels using a subset of Target2 data generated as part of the the JUMP-CP effort\cite{jump_target2} to predict Cell Painting \cite{bray2016} images from paired brightfield images \cite{CrossZamirskiGAN}. We find that the quality of extracted morphological features from the predicted images, and their performance on downstream mechanism of action prediction and clustering tasks can be boosted with relevant labels. This type of approach may lead to increased clinical success of image-to-image methods in drug discovery \cite{Chandrasekaran2021} and related medical reconstruction tasks \cite{Yang2022}, as a way to guide the image generation with biologically informative class information. 
In this study we make the following contributions:
\begin{itemize}
    \item We introduce and implement a general framework for class-guided image-to-image diffusion, our model building upon the \textit{Palette} image-to-image framework \cite{Saharia2021_palette} and guided diffusion \cite{Dhariwal2021}  .
    \item We apply our model to the prediction of 5-channel Cell Painting fluorescent microscopy from 3-channel brightfield images, and show that incorporating label information can improve performance. We evaluate the images with extracted biological features and a transfer learning approach to simulate image-based profiling in a drug discovery pipeline.
\end{itemize}


\section{Related work}

Generative adversarial networks (GANs) \cite{goodfellow2014} have been the prevailing method for image-to-image translation tasks since the introduction of \textit{pix2pix} \cite{isola2017} in 2016. GAN based methods have been widely adopted in medical imaging for a variety of tasks \cite{Yi2019} including PET denoising, PET-CT translation and correction of magnetic resonance motion artefacts \cite{Armanious2020}. GANs are used in cell microscopy for cross-modality prediction \cite{Belthangady2019} and super resolution \cite{Zhang2019}.  

Other models used for reconstruction tasks include variational auto-encoders (VAEs) \cite{Kingma2013} and normalizing flows \cite{Kingma2018}. VAEs have been used to learn or approximate the joint distribution of multiple modalities \cite{Wu2018}, sometimes with a product or mixture of experts approach to combine the distributions \cite{Shi2019}. Product of experts have also be used for multimodal conditional image synthesis with GANs \cite{Huang2022}. Flow-based models for modality transfer (such as MRI to PET) have outperformed conditional GANs and VAEs while leveraging \textit{side} information \cite{Sun2019}.

Diffusion models are growing in popularity in medical imaging and have been used predominantly for MRI and CT modalities in reconstruction problems \cite{Kazerouni2022}. Diffusion-based generative models can achieve state of the art image quality without suffering from problems such as mode collapse, training instability, or not allowing for likelihood estimation. By comparison, GANs can suffer from training instability and mode collapse \cite{Creswell2018} in addition to feature hallucinations which are particularly undesirable in medical applications \cite{Cohen2018}. VAEs do not produce high quality image samples and flow-based models have restrictions such as the requirement for invertibility of the network. 

Weakly-supervised diffusion models have been used in medical imaging, notably in anomaly detection \cite{Sanchez2022, Wolleb2022}. However, these models are not strictly guided image-to-image models and instead use the difference between the ground truth and reconstructed image for anomaly detection. This method would not generalize beyond anomaly detection. \textit{InstructPix2Pix} \cite{Brooks2022} combines a text-guided conditional diffusion model with an image-to-image framework using text based prompts. However, text encoders for style-transfer are not appropriate in datasets where metadata labels are discrete classes.


Hence there is scope for developing a diffusion model for image reconstruction with discrete metadata. Examples of image-to-image problems with (often under-utilised) data include: prediction of fluorescent image channels from transmitted light images for drug discovery \cite{christiansen2018} where freely available weak labels include treatment and compound \cite{caicedo2018}, as well as batch information. Experimental batch effects can be significant, and batch information has been integrated into a number of machine learning models in image-based profiling \cite{Lin2022, Wang2021, ando2017}. PET reconstruction from much cheaper MRI scans for Alzheimer's prediction also has inexpensive and relevant metadata such as patient age, sex, disease status and genotype which has been incorporated into improving image reconstruction quality \cite{Sun2019}.


To the best of the authors' knowledge, our work is the first to use a diffusion model for fluorescent microscopy prediction. We build upon existing studies using deep learning to predict fluorescent labels \cite{christiansen2018} from transmitted light images such as brightfield, a cheaper and less invasive modality for imaging cells which can still capture meaningful information \cite{gupta2022}. Specifically, we predict Cell Painting \cite{bray2016} image channels which capture rich cell morphology information which can be used in a variety of tasks in image-based profiling including bioactivity, cytotoxicity and mechanism of action prediction \cite{Chandrasekaran2021}.

\subsection{Image-to-image conditional diffusion}

We base our model on the \textit{Palette} framework for image-to-image diffusion from Saharia \textit{et al.} \cite{Saharia2021_palette}. Their model outperforms GANs on four tasks: colorization, inpainting, uncropping and JPEG restoration. \textit{Palette} is a denoising diffusion probabilistic model \cite{Ho2020} of the form $p(\bm{y} \mid \bm{x})$ which is trained to predict the output image $\bm{y}$ conditional on the input image $\bm{x}$. The noisy image $\widetilde{\bm{y}}$ is given by:
\begin{equation} \widetilde{\bm{y}} = \sqrt{\gamma} \bm{y} + \sqrt{1-\gamma} \bm{\epsilon}, \quad \bm{\epsilon}\sim\mathcal{N}(\bm{0},\bm{I}).\end{equation} 
for Gaussian noise $\bm{\epsilon} \sim\mathcal{N}(\bm{0},\bm{I})$ and noise level indicator $\gamma$. A neural network $f_\theta$ is trained to denoise $\widetilde{\bm{y}}$ for a given $\bm{x}$ with the loss function: 
\begin{equation}\label{Ho} \mathbb{E}_{(\bm{x},\bm{y})} \mathbb{E}_{\bm{\epsilon}\sim\mathcal{N}(\bm{0},\bm{I})} \mathbb{E}_\gamma \bigg\lVert f_\theta(\bm{x}, \underbrace{\sqrt{\gamma} \bm{y} + \sqrt{1-\gamma} \bm{\epsilon}}_{\widetilde{\bm{y}}}, \gamma) - \bm{\epsilon} \bigg\rVert^{p}_p~ \end{equation} 
where $p$ is the chosen norm ($L_1$ or $L_2$). Eq. \eqref{Ho} is the image-conditional version of $L_{simple}$ from Ho \textit{et al.} \cite{Ho2020}. 

The reverse diffusion process is computed step-by-step as:
\begin{equation}
\bm{y}_{t-1} \leftarrow \frac{1}{\sqrt{\alpha_t}} \left(\bm{y}_t - \frac{1-\alpha_t}{ \sqrt{1 - \gamma_t}} f_{\theta}(\bm{x}, \bm{y}_{t}, \gamma_t) \right) + \sqrt{1 - \alpha_t}\bm{\epsilon}_t
\end{equation}
for $t = T,\dotsc,1$ steps. The noise level indicator $\gamma_t$ is a function of $t$, and $\alpha_t$ is the noise variance scale parameter (also timestep-dependent).  
%

\subsection{Conditional image synthesis}

For conditional image synthesis with class labels, Dhariwal and Nichol \cite{Dhariwal2021} introduced two modifications to unconditional DDPM from Ho \textit{et al.} \cite{Ho2020}: adaptive group normalization (AdaGN) and classifier guidance (CG). AdaGN is a modification to the architecture which incorporates the class information into normalization layers in training, while classifier guidance exploits the gradients of a pre-trained classifier to guide the inference process (note: in this section we change $y \rightarrow k$ and $\bm{x}\rightarrow \bm{y}$ from the original paper to be consistent with the notation used in this paper)

\subsubsection{Adaptive group normalization}

AdaGN is a layer used to incorporate the timestep and class embedding into the residual blocks following a group normalization operation \cite{Wu2018_GN}. It is defined as: \begin{equation}
\text{AdaGN}(h,k) = k_s \text{GroupNorm}(h) + k_b
\end{equation}
where $k = [k_s,k_b]$ is a linear projection of the timestep and class embedding, and $h$ is the activations of the residual block after the first convolution. This layer can be incorporated in the absence of class labels with just the timestep embedding: AdaGN $= k_s\text{GroupNorm}(h)$.

\subsubsection{Classifier guidance}

Classifier guidance enables the use of class information in inference of the trained diffusion model. Sohl-Dickstein \textit{et al.} \cite{SohlDickstein2015} and Song \textit{et al.} \cite{Song2020} showed this can be achieved using pre-trained classifier gradients to condition the sampling of the diffusion model. First, the classifier $p_\phi(k\mid \bm{y}_t)$ is pre-trained to predict the class $k$ from noisy images $\bm{y}_t$. 

The aim is to sample each transition from the distribution:
\begin{equation}\label{intractable}
p_{\theta,\phi}(\bm{y}_t \mid \bm{y}_{t+1},k) = Zp_{\theta}(\bm{y}_t \mid \bm{y}_{t+1})p_{\phi}(k \mid \bm{y}_t)
\end{equation}
where $p_\theta(\bm{y}_t\mid \bm{y}_{t+1})$ is the unconditional reverse noising process and $Z$ is a normalizing constant. Although it is intractable to sample from the distribution in Eq. \eqref{intractable}, it can be approximated as a perturbed Gaussian distribution \cite{SohlDickstein2015}:
\begin{equation}
\log (p_{\theta}(\bm{y}_t \mid \bm{y}_{t+1})p_{\phi}(k \mid \bm{y}_t)) \approx \log p(\bm{z}) + C, 
\end{equation}
\begin{equation}
\bm{z}\sim\mathcal{N}(\bm{\mu} + \bm{\Sigma} g,\bm{\Sigma}), \quad g = \nabla_{\bm{y}_t}\log p_{\phi}(k\mid \bm{y}_t)|_{\bm{y}_t = \mu}
\end{equation}
where $g = \nabla_{\bm{y}_t}\log p_{\phi}(k\mid \bm{y}_t)$ are the gradients of the classifier and $C$ is a constant which can be ignored. In inference, this shifts the mean of the sampled Gaussian to guide the denoising process towards the given class label $k$. The relative weighting of the classifier guidance term can be scaled with a constant $s$.

\section{Class-guided image-to-image diffusion}

In \textit{Palette}, Saharia \textit{et al.} \cite{Saharia2021_palette} removed both classifier guidance and the class embedding of the AdaGN layer introduced by Dhariwal and Nichol \cite{Dhariwal2021}. In this study we re-introduce the class label $k$ while retaining the conditional dependence on input image $\bm{x}$. We redefine the input conditions for image and associated class label as $(\bm{x}_k, k)$.


\begin{algorithm}
  \caption{Training the denoising model $f_\theta$}
  \begin{algorithmic}
    \REPEAT
      \STATE $(\bm{x}_k, \bm{y}_0, k) \sim p(\bm{x}_k, \bm{y}, k)$
      \STATE $\gamma \sim p(\gamma)$
      \STATE $\bm{\epsilon}\sim\mathcal{N}(\bm{0},\bm{I})$
      \STATE Take a gradient descent step on \\
      $\qquad \nabla_\theta \left\lVert f_\theta(\bm{x}_k, \sqrt{\gamma} \bm{y}_0 + \sqrt{1-\gamma} \bm{\epsilon}, k, \gamma) - \bm{\epsilon} \right\rVert_p^p$ 
    \UNTIL{converged}
  \end{algorithmic}
\end{algorithm}

\begin{algorithm}
\caption{Classifier guided diffusion sampling, given a diffusion model ($\bm{\mu}_{\theta}(\bm{x}_t), \bm{\Sigma}_{\theta}(\bm{x}_t))$, classifier $p_{\phi}(k\medspace|\medspace \bm{y}_t)$, and gradient scale s.}\label{alg:sampling}

  \begin{algorithmic}[1]
    \STATE Input: class label $k$, input image $\bm{x}_k$, gradient scale $s$
    \STATE $\bm{y}_T \sim \mathcal{N}(\bm{0}, \bm{I})$
    \FOR{$t=T, \dotsc, 1$}
    \STATE  $\bm{z} \sim \mathcal{N}(\bm{\mu} + s \Sigma\nabla_{\bm{y}_t}\log p_{\phi}(k\mid \bm{y}_t), \bm{\Sigma})$ if $t>1$, \\ else $\bm{z} = \bm{0}$
   \STATE  $\bm{y}_{t-1} = \frac{1}{\sqrt{\alpha_t}}\left(\bm{y}_t - \frac{1-\alpha_t}{\sqrt{1-\gamma_t}} f_\theta(\bm{x}_k, \bm{y}_t, k, \gamma_t) \right) + \sqrt{1 - \alpha_t} \bm{z}$
    \ENDFOR
    \STATE \textbf{return} $\bm{y}_0$
  \end{algorithmic}
\end{algorithm}

We summarise the training scheme for class guided image-to-image diffusion in Algorithm 1, and the sampling scheme in Algorithm 2. Each iteration of the reverse process of class-guided image-to-image diffusion can be computed as:
\begin{multline}
    \bm{y}_{t-1} = \frac{1}{\sqrt{\alpha_t}}\left(\bm{y}_t - \frac{1-\alpha_t}{\sqrt{1-\gamma_t}} f_\theta(\bm{x}_k, \bm{y}_t, k, \gamma_t) \right) \\ + \sqrt{1 - \alpha_t} \bm{z}
\end{multline}
for $t = T,\dots,1$. Here:
\begin{multline}
    \bm{z} \sim \mathcal{N}(\bm{\mu} + s \bm{\Sigma}\nabla_{\bm{y}_t}\log p_{\phi}(k\mid \bm{y}_t), \bm{\Sigma}) \quad \text{if} \quad  t>1, \\ \quad \text{else} \quad \bm{z} = \bm{0}
\end{multline}
The $f_\theta$ dependence on $k$ is achieved with the AdaGN layer. It is optional to use $k$ in sampling, as the AdaGN layer does not need to see the label. We test this and find that although it is possible to exclude $k$ in sampling, it is necessary to include for improved performance. It is also possible to sample without classifier guidance by setting $s=0$. 

The training objective follows the form of  Eq. \eqref{Ho} with $L_2$ norm. Our network is a U-Net architecture \cite{Ho2020} which is based on the modified  $256 \times 256$ class-conditional U-Net model used in \textit{Palette} \cite{Dhariwal2021, Song2021}. The network is adapted to take images of size $512 \times 512$ with 3 input channels and 5 output channels in order to fit the requirements of the brightfield and Cell Painting channels.

\section{Experiments}
\subsection{Dataset}


We used a subset of one of the publicly-available JUMP Cell Painting dataset cpg0000 \cite{chandrasekaran2022a}, available from the Cell Painting Gallery on the Registry of Open Data on AWS (https://registry.opendata.aws/cellpainting-gallery/). 10 plates (experimental replicates) were chosen to ensure a variety of biological phenotypes were present. They contain pairs of compounds associated by the genes they target, in addition to 46 controls compounds with a variety of mechanisms. In total, every plate contains around 2000 images - each with 5 Cell Painting channels and 3 brightfield channels. These plates were screened regularly throughout data production to enable downstream assessment of connectivity of perturbations between batches of compounds screen. Every plate contains treated cells representing 290 perturbations, each with paired perturbation with a matching target (145 targets total). Imaging details are provided in the supplementary material.

\subsection{Pre-processing}

To ensure that systematic variations in pixel intensity were not present in input images, we used a standardised CellProfiler \cite{carpenter2006} pipeline to perform illumination correction on all images. A smoothing function of filter size 249 pixels was used to generate an illumination correction function per imaging channel for each plate. The pixel intensities of all images were then divided by their respective correction function. This methodology is consistent with best practice established during the JUMP Cell Painting consortium \cite{cimini2022}. After illumination correction, all images were re-sized to $512 \times 512$ pixels using bicubic interpolation. The images were all normalised to have a standard deviation of 1 and a mean of 0 with a maximum pixel intensity cutoff of 15 enforced to exclude extreme outliers.

\subsection{Model training}

We trained each model using 9 training plates and evaluated on a single, unseen test plate. For each model this was done twice, learning weights for 2 different, randomly selected test plates (the same 2 plates for each model). This is equivalent to k-fold cross validation - although we trained 2 versions of each model rather than 10, as producing 10 full plates per model was not possible due to the computationally intensive nature of sampling DDPMs. 

Using the full plates, we trained models with no labels (\textit{Palette}), perturbation (pert) as a weak label, target as a label. The labels were included through the AdaGN layer. We compared using the labels in training and inference against using labels in training but not in inference through the AdaGN layer to test if the labels were required in sampling. Target as a label was included as a proof of concept of the method, but it is expensive information not freely available in a practical setting (compared to the perturbation which is free information). Classifier guidance was not used to generate entire plates due to the extreme computational demands (over 500 GPU hours per model, per plate). This training regime is equivalent to Algorithm 2 with $s=0$. 

The active subset was around one third of the full plate, and this allowed us to sample using classifier guidance with $s=1$. Additionally, training with known active compounds would provide more meaningful class labels for the model. The classifier $p_\phi(k\mid \bm{y}_t)$ was the downsampling branch of the U-Net with an additional output layer (as inroduced by Ho \textit{et al.} \cite{Ho2020}). Training images were noised with the timestep dependent noise distribution, and the model was trained until the loss converged. The full training and sampling schemes are presented in Algorithms 1 and 2.

In training, the images were subject to random horizontal and vertical flips and 90 degree rotations each with probability $p=0.5$. Models were trained until the loss appeared to stop decreasing, which was typically around $250,000$ iterations. Even though the quality of cellular structures appeared to improve beyond this, we found overfitting to be a problem for larger number of epochs as phantom structures appeared on the empty background. All models were trained with a batch size of 2 and the Adam optimizer with a learning rate of $8e^{-5}$. The linear noise schedule of $(10e^{-6}, 0.001)$ (as in \textit{Palette}) with $T=2000$ was used in training and inference. We provide the code and parameters to replicate these models in our GitHub repository. All the models were trained on the AstraZeneca Scientific Computing Platform (SCP) with 32GB GPUs. Total training time was around 24 hours on a single GPU, and sampling from the trained model was around 4 minutes per 5 channel image (increasing to 15 minutes with classifier guidance).

\subsection{Post-processing}

The model outputted channels were re-normalised as in the pre-processing. CellProfiler \cite{carpenter2006} was used to segment nuclei, cells and cytoplasm, then extract morphological features from each of the channels. Single cell measurements of fluorescence intensity, texture, granularity, density, location and various other features were calculated as feature vectors. Features were aggregated for each perturbation using the median value per image. 

The Pycytominer package (\url{https://github.com/cytomining/pycytominer})  was used to normalise the cell-painting features generated for the synthetic images. The features derived for synthetic images generated by each model were normalized using all the samples. All the features generated from the  ground truth data were also used for the prediction feature selection operation to allow for a fair comparison. These included dropping na columns, variance thresholding, correlation thresholding and dropping blocklisted features. Approximately 650-700 cell-painting features were selected for each plate. Features were aggregated to the perturbation level, giving 290 features per plate.

In order to segregate the active perturbations from inactives, PCA was performed using 1262 Cell Painting features that remained after CellProfiler feature selection (from the ground truth images). The top 100 dimensions of the PCA were then used to evaluate the cosine-distances between all-pairs of data points (well). An average cosine-distance score against the negative DMSO controls across all replicates was used as a score to segregate out the actives from inactives using 1D C-kmeans clustering algorithm with $k=3$ for 3 clusters. There were a total of 118 perturbations representing 59 targets selected for the active subset, with the remaining perturbations (inactives) showing no phenotypic divergence from negative controls. We provide visualisations of the dataset and the active subset in the supplementary material.

\subsection{Transfer learning with DINO}

We used the self-supervised learning algorithm DINO \cite{caron2021} pre-trained with ImageNet \cite{deng2009} weights to profile the images with transfer learning, following the methodology of previous studies \cite{CrossZamirskiWSDINO}. The backbone of the network is a vision transformer (ViT-S/8) with a 3-layer multi-layer perceptron head, from which the embeddings are extracted. The median feature embedding was taken from four $224 \times 224$ crops around the centre of each image (equivalent to a $448 \times 448$ pixel centre crop split into four non-overlapping crops). Embeddings of size 384 were extracted for each channel then concatenated to a size of 1920 for 5-channel Cell Painting (1152 for brightfield) followed by $L_2$ normalization. These feature representations, like the CellProfiler features, were then used for target prediction.

\section{Results}

\subsection{Evaluation}

\begin{table*}[h]
\centering
    \small
    {
    \begin{tabular}{cccccccccc}
    
    \toprule
    \multicolumn{2}{c}{\textbf{Label}}  & \multirow{2}{*} {\bfseries{PCC} $\uparrow$}& \multirow{2}{*}{\bfseries{FID} $\downarrow$}& \multirow{2}{*}{\bfseries{SSIM} $\uparrow$}& \multirow{2}{*}{\bfseries{MSE / MAE} $\downarrow$} & \bfseries{NN matches $\uparrow$}  &  \bfseries{NN Top 5 $\uparrow$}  & \bfseries{MTdist $\downarrow$} & \multirow{2}{*}{\bfseries{CPcor} $\uparrow$} \\ 
    \bfseries{Training} & \bfseries{Sampling} & & & & & \textbf{CP / TL} & \textbf{CP / TL} & \textbf{CP / TL}\\
    \midrule
    None & None  & \textbf{0.793} & 3.54 & \textbf{0.350} & \textbf{0.400} / \textbf{0.338} &  6 / \textbf{8} &  \textbf{23} / \textbf{25} & \textbf{0.886} / \textbf{0.0729} & \textbf{0.430} \\
    Pert & None  & 0.760 & \textbf{3.26} & 0.267 & 0.465 / 0.380  & \textbf{7} / 5 & 20 / \textbf{25} & 0.910 / 0.0852 & 0.384 \\
    Pert & Pert  & 0.752 & 3.49 & 0.260 & 0.481 / 0.392  & \textbf{7} / 3 &  22 / 22 & 0.919 / 0.0936 & 0.381  \\
    \midrule
    Target$^*$ & None  & 0.741 & 3.69 & 0.239 & 0.489 / 0.402 & 4 / 4 & 15 / 17 & 0.888 / 0.0834 & 0.283   \\
    Target$^*$ & Target$^*$  & 0.745 & 3.86 & 0.228 & 0.495 / 0.408 & \textbf{17} / \textbf{15} &  3\textbf{2} / \textbf{51} & \textbf{0.791} / 0.0836 & 0.327   \\
    \midrule
    \multicolumn{2}{c}{GT Cell Painting}  & $-$ & 1.55\textsuperscript{\textdagger} & $-$ & $-$ &  12 / 13 & 31 / 28 & 0.868 / 0.0924 & 0.569\textsuperscript{\textdagger} \\
    \midrule
    \multicolumn{2}{c}{GT Brightfield}  & $-$ & $-$ & $-$ & $-$ & $-$ / 12 & $-$ / 28 & $-$ / (0.0551) & $-$\\
    \bottomrule
    \end{tabular}
    }
    \caption{Mean image and feature metrics for class-guided image-to-image models for two full plates, each generated with a different model. Note the brightfield feature space (3 channels) is a different size to the Cell Painting feature space (5 channels). $^*$Target is not a freely available label and is included as a proof of concept.  \textsuperscript{\textdagger}We provide FID and CPcor values calculated between the two ground truth (GT) test plates, which are prepared and treated as identical replicates.}
    \label{tab:tab1}
\end{table*}

\begin{table*}[h]
\centering
    {
    \begin{tabular}{ccccccccc}
    
    \toprule
    \multicolumn{2}{c}{\textbf{Label}}  & \multirow{2}{*} {\bfseries{PCC} $\uparrow$} & \multirow{2}{*}{\bfseries{SSIM} $\uparrow$}& \multirow{2}{*}{\bfseries{MSE / MAE} $\downarrow$} & \bfseries{NN matches $\uparrow$}  &  \bfseries{NN Top 5 $\uparrow$}  & \bfseries{MTdist $\downarrow$} & \multirow{2}{*}{\bfseries{CPcor} $\uparrow$} \\ 
    \bfseries{AdaGN} & \bfseries{CG} & & & & \textbf{CP / TL} & \textbf{CP / TL} & \textbf{CP / TL}\\
    \midrule
    None & None  & \textbf{0.773} &  0.294 & \textbf{0.423} / 0.320  & 4 / 2  & 12 /  16 & 0.971 / 1.533 & 0.386 \\
    Pert & None  & 0.762 &  \textbf{0.379} & 0.444 / \textbf{0.310}  & 6 / 4 &  18 / \textbf{18} & 0.939 / \textbf{1.357} & \textbf{0.507} \\
    Pert & Pert  & 0.752 &  0.338 & 0.463 / 0.330   & \textbf{7} / \textbf{7} & \textbf{24} / \textbf{18} & \textbf{0.929} / 1.541 & 0.504 \\
    \midrule
    Target$^*$ & None  & 0.730 &  0.235 & 0.506 / 0.375 & 9 / 14 & 27 / \textbf{27} & 0.883 / 1.405 & 0.404 \\
    Target$^*$ & Target$^*$ & 0.696 & 0.202 & 0.573 / 0.408  & \textbf{11} / 6 & \textbf{31} / 21 & \textbf{0.879} / 1.579 & 0.355 \\
    \midrule
    \multicolumn{2}{c}{GT Cell Painting}  & $-$ & $-$ & $-$  & 9 / 13 & 21 / 26 & 0.919 / 0.233 & 0.615\textsuperscript{\textdagger} \\
    \midrule
    \multicolumn{2}{c}{GT Brightfield}  & $-$ & $-$ & $-$ & $-$ / 16 & $-$ / 26 & $-$ / (1.148) & $-$ \\
    \bottomrule
    \end{tabular}
    }
    \caption{The analysis of Table 1 is repeated for the active subset only. There are too few images in the active subset to calculate FID. $^*$Target is not a freely available label and is included as a proof of concept. \textsuperscript{\textdagger}We provide the CPcor value calculated between actives in the two ground truth (GT) test plates, which are prepared and treated as identical replicates.}
    \label{tab:tab2}
\end{table*}

\begin{figure*}
  \centering
  \includegraphics[width=1\textwidth]{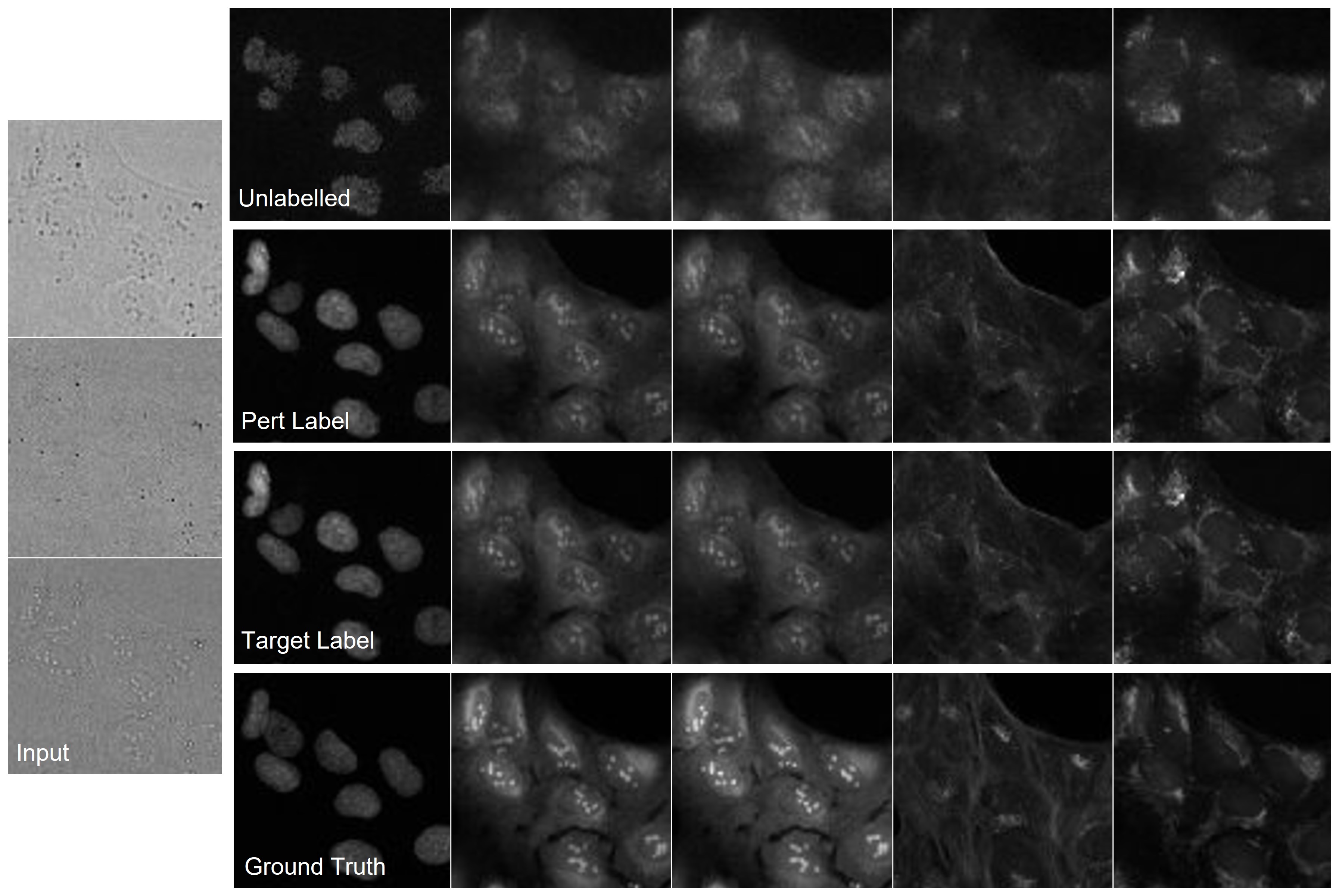}
  \caption[fig3]{Images generated by models trained with the active subset. The labelled images were sampled with both AdaGN and CG. Cropped to $100 \times 100$ pixels. Columns left to right: Brightfield (input), DNA, RNA, ER, Mito, AGP.}
  \label{lch616}
\end{figure*}

We evaluated our models with image-level and feature-level metrics, which are presented in Table \ref{tab:tab1} (the entire plate) and Table \ref{tab:tab2} (the active subset). We compared Pearson correlation coefficient (PCC), Fréchet Inception Distance (FID) \cite{Heusel2017}, structural similarity index measure (SSIM) \cite{hore2010} and mean-squared and mean-absolute error (MSE/MAE). The values in the tables were calculated by comparing the predicted images with the ground truth images for each model. The values presented are the mean values of all the images. Examples of the images are presented in Figures \ref{fig:fig1}, \ref{lch61} and \ref{lch616}. We also compare the FID scores and feature values between the two ground truth plates as the limit of a perfect reconstruction (each plate is meant to be an experimental replication of the same cells and treatments). The feature-level metrics were chosen to be representative of downstream applications which would be performed with real Cell Painting images in a drug discovery pipeline.

\subsubsection{NN matching / NN top 5}

We searched the feature spaces of each plate - both CellProfiler (CP) and transfer learning (TL) spaces - for the nearest neighbours by cosine distance. The values reported in the tables are the total number of matching targets which are nearest neighbours in the feature space of the model or ground truth plate feature space (for both plates). We repeated this analysis but for each point searching for the 5 nearest neighbours, and reporting a match if one of the 5 perturbations shared a target with the chosen point. 

\subsubsection{Matching target distance (MTdist)}

Since there hundreds of targets and perturbations in this dataset, even searching the top 5 nearest neighbours is not sufficient to evaluate the relationships between targets. We propose the mean matching target distance (MTdist) as an informative metric. For each pair of perturbations sharing a matching target, the cosine distance is calculated between the points in feature space. The mean distance for all 290 perturbations (118 in the active subset) is presented for each model. Since the models feature spaces are normalised this should be a fair comparison between the models. 

\subsubsection{CellProfiler feature correlation (CPcor)}

Following the methodology of previous Cell Painting prediction studies \cite{CrossZamirskiGAN}, we correlated each model's CellProfiler features to the ground truth CellProfiler features, and report the mean value. We also correlate the features between the ground truth replicates as a baseline (0.569 for the whole plate and 0.615 for the active subset). We would not expect the model generating features from an unseen batch to exceed this value. We include a breakdown of the features by group and channel in the supplementary material, alongside two-dimensional t-SNE plots of the features.

\section{Discussion and conclusion}

The purpose of this study was to explore how metadata in the form of discrete classes can be used to guide image-to-image translation tasks. DDPMs and other generative models have been successful in achieving state of the art FID scores, however learning details which differentiate between images based on biology and structure is less studied when compared to generating realistic images which could have been sampled from a training distribution. 

All the models achieved very low FID scores. For entire plates (Table \ref{tab:tab1}), incorporating labels through AdaGN generally reduced the performance of the pixel-level metrics, although using the perturbation as a label in training resulted in the lowest FID score. Some images from the labelled models had some background noise which was not present in the unlabelled model, and this is reflected in the image-level metrics. We present an example of this effect in the supplementary material. While it is possible that class labels can improve certain aspects of the images, they may also reduce the image quality by fitting to unwanted background noise if there are uninformative class labels or no signal to be found in the training set. This was particularly notable using target as the label, which moved matching targets closer in feature space, but reduced the faithfulness of the generated image. This effect is likely amplified in high-content microscopy images where over 50\% of the pixels are irrelevant background with no cellular structures. 

The results for the model trained with full plates of images (290 perturbations) suggest that the image-to-image model is capable of capturing strong phenotypic signals (true positives) but struggled with noisy, lower signal images (the inactives). We may have led the model astray with uninformative labels in training. To test this theory, we repeated the analysis with the active subset (Table \ref{tab:tab2}), which represents the images of cells with meaningful and quantifiable phenotypic differences from the control group (untreated cells). This resulted in a significant improvement over the unlabelled model (\textit{Palette}) in SSIM, target matching and CellProfiler feature correlation. Furthermore, the pixel correlations and errors were not significantly reduced, so unlike when using the whole plate, there was less of a cost to the improved performance. Incorporating classifier guidance improved target matching but also at a small cost to image and feature quality. Our results show that class-guided image-to-image diffusion improves upon the naive model under well-chosen conditions, and highlight how crucial the quality of labels and training data is.

The values in Tables \ref{tab:tab1} were produced by models trained and tested on the whole plate, while Table \ref{tab:tab2} presents results from images trained with the smaller active subset. The smaller training set of the active subset reduced the quality of the unlabelled model. However, incorporating labels produced the highest correlations of features, and the highest SSIM even in the low training data regime. These effects were observed in both training splits. Very recently, Cell Painting datasets of immense scale with millions of images across thousands of compounds and over 50 batches have become public, and hold great promise for machine learning in drug discovery \cite{Sypetkowski2023, Fay2023}. The batch effect is a large part of this, and we explore the batch effect properties of our models in the supplementary material.

This study provides a valuable comparison of methods employing brightfield image channels as an input for image-based profiling. Recent studies have explored this under-utilised modality which may contain as much predictive power as fluorescent stained images \cite{gupta2022}. Our results further reveal the potential of brightfield both as an input for cross modality prediction and as a competitive profiling modality in itself. This success may also be attributed to powerful, pretrained attention based architectures \cite{caron2021}, which can overcome the traditional drawbacks of brightfield and are able to find meaningful structures from noisy images (we present self-attention maps of transfer learning with brightfield images in the supplementary material). Brightfield and transmitted light has traditionally been seen as less informative than fluorescent staining, but the limit of brightfield may be higher than previously thought. Furthermore, we have presented a way to use brightfield to generate full plates of model-generated Cell Painting, from which existing software can extract hand-crafted features for a greater level of interpretability. 

In conclusion, we present a novel way to use discrete metadata to guide image-to-image translation. We predict unseen batches of Cell Painting from brightfield, and surpass the performance of previous methods in multiple metrics \cite{CrossZamirskiGAN}. We perform image-based profiling predictions with the model predicted plates and achieve stronger results when using the freely available perturbation label with the active subset. This includes phenotypic feature correlations, SSIM and target matching, a common task in drug discovery. We propose our method could have impact in other biomedical fields to guide learning meaningful features and structures with multimodal data.

\clearpage

\section*{Supplementary Material}

\appendix       

\section{Dataset}

\subsection{Further imaging details}

In the Cell Painting assay cell phenotypes are captured with six generic fluorescent dyes and imaged across five channels. The assay is designed to visualise eight cellular components: nucleus (DNA channel), endoplasmic reticulum (ER channel), nucleoli, cytoplasmic RNA (RNA channel), actin, Golgi, plasma membrane (AGP channel) and mitochondria (Mito channel) \cite{bray2016}.

For the data in this study U2-OS cells were incubated in 5µM compounds for 48h, then fixed and stained according to the updated Cell Painting protocol \cite{cimini2022}. Plates were imaged on a CellVoyager CV8000 (Yokogawa) with a water-immersion $20\times$ objective (NA 1.0). Excitation and emission wavelengths were as follows for fluorescent channels: DNA (ex: 405nm, em: 445/45nm), ER (ex: 488nm, em: 525/50nm), RNA (ex: 488nm, em: 600/37nm), AGP (ex: 561nm, em: 600/37nm) and Mito (ex: 640nm, em: 676/29nm). The three brightfield images were acquired from different focal z-planes; within, 4µm above and 4µm below the focal plane. Images were saved as 16-bit .tiff files with $2\times2$ binning ($998 \times 998$ pixels).

\subsection{Active subset selection}

\begin{figure*}
  \centering
  \includegraphics[width=0.88\textwidth]{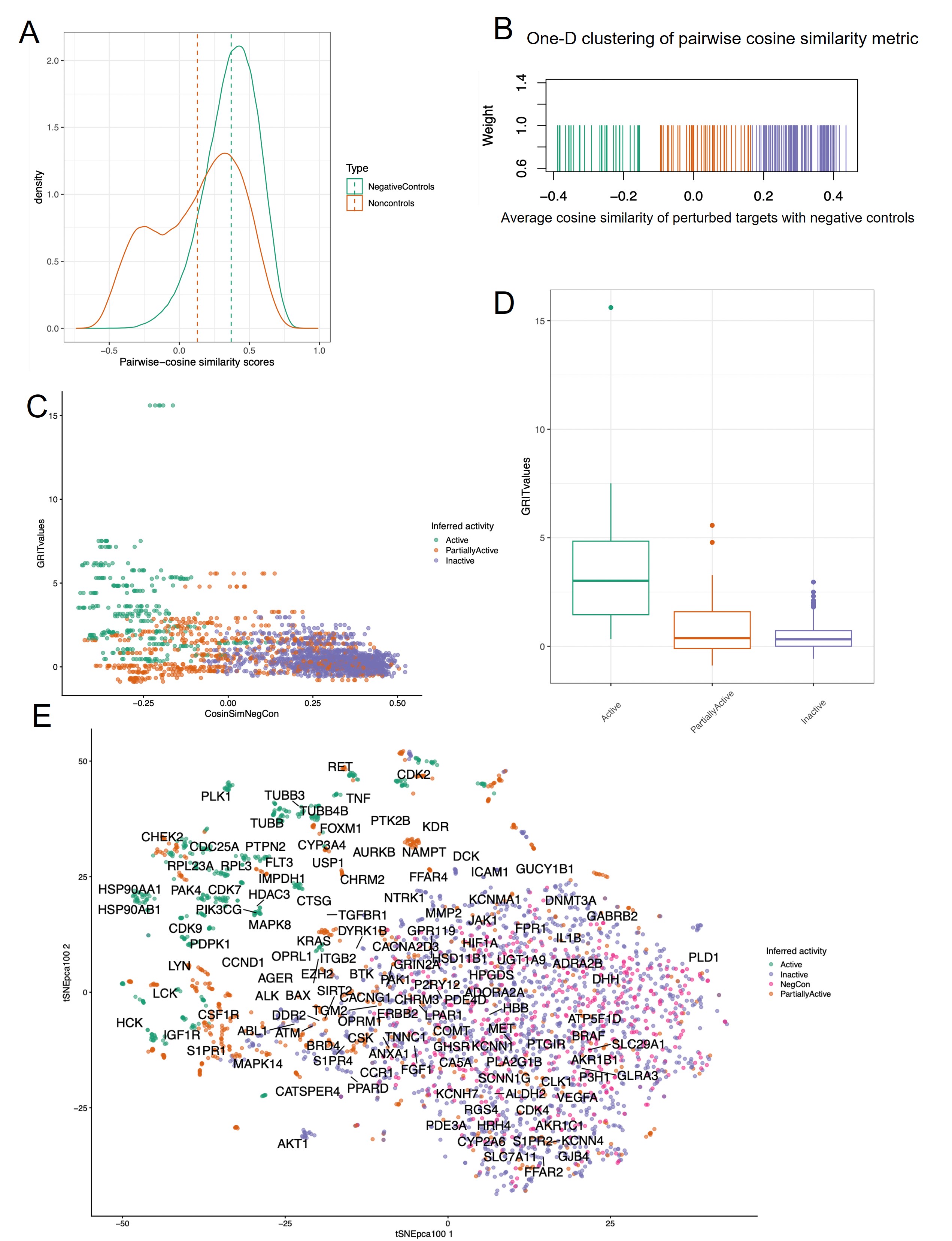}
  \caption[s]{Inferring the target activity with the ground truth Cell Painting CellProfiler features. \textbf{A.} The distribution of all the pairwise cosine similarity scores derived from the top 100 PCA dimensions across the negative controls and the drug treatments. \textbf{B.} One-dimensional K-means clustering of the average cosine similarity metric computed between the targets and negative controls. \textbf{C.} Scatter plot of the Grit values computed for each target and the corresponding cosine similarity metric calculated from the negative controls. \textbf{D.} Box plot depicting Grit values across inferred target activity. \textbf{E.} Two-dimensional t-SNE plot of all the 10 TARGET-2 plates colored based on the inferred target activity.}
  \label{lch6612}
\end{figure*}

In Fig. \ref{lch6612} we provide an overview of how the active subset was selected. We identified three groups of treated cells based on their CellProfiler features across all 10 plates: the active, partially active and inactive groups. The inactive subset overlapped with the negative control (DMSO) subset. Fig. \ref{lch6612}E provides an intuitive visualisation of our choice to train the model with the active subset only. Our results using the entire plate show that using labels which do not correspond to structural and biological differences will reduce the generated image quality. We provide examples of lower quality images in Section \ref{sec:appendixC}.

Although we do not use it for selecting our subsets, we present a visualisation of Grit scores for our dataset in Fig. \ref{lch6612}. Grit is a calculation used in image-based-prifling to define how different a perturbation or compound is from the DMSO controls (\url{https://github.com/broadinstitute/grit-benchmark}).

\section{Further results and figures}\label{sec:appendixC}


\subsection{Cell Painting feature breakdown}

\begin{figure*}
  \centering
  \includegraphics[width=1\textwidth]{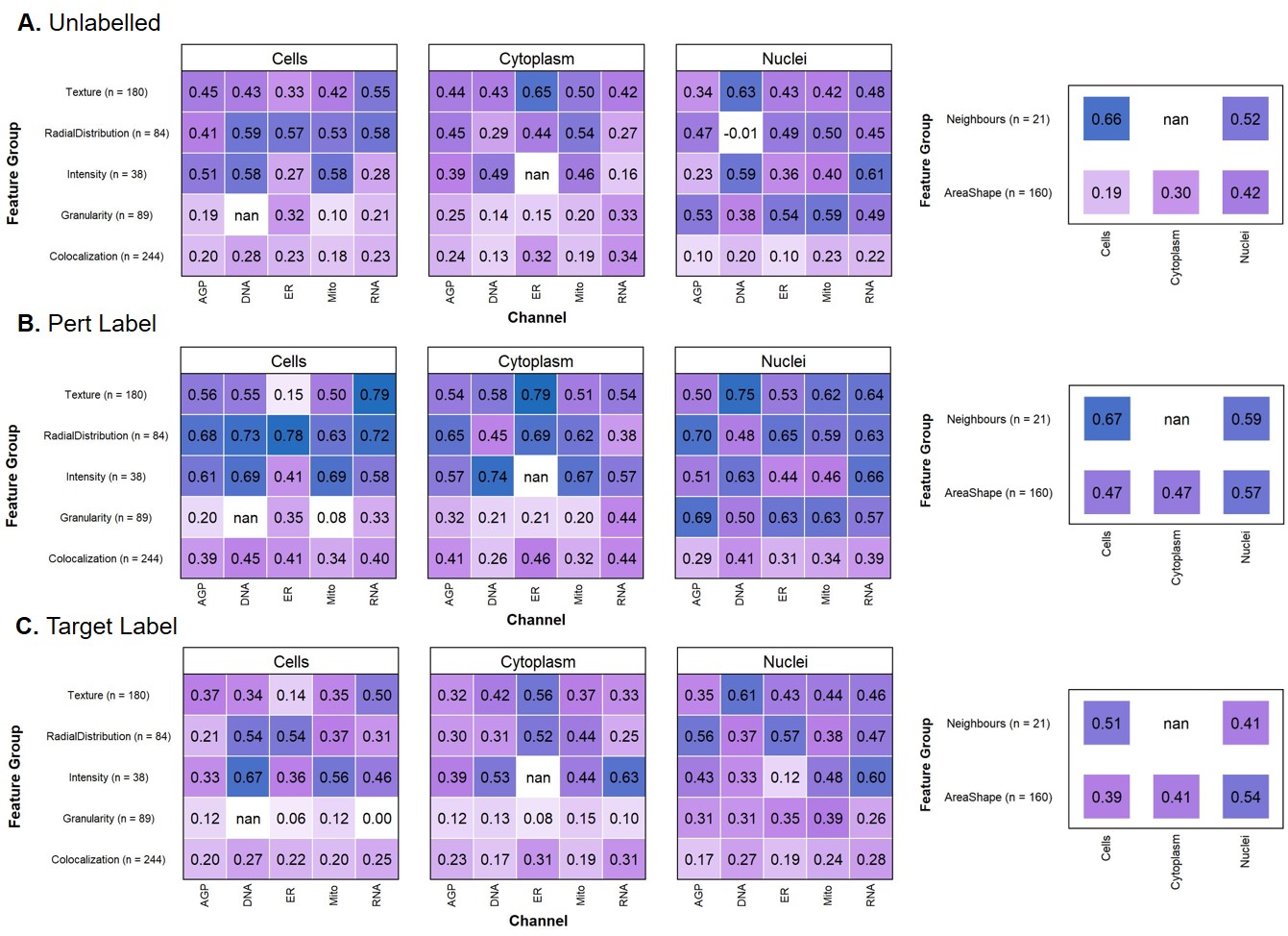}
  \caption[sc23]{Heatmaps of the mean correlations to the ground truth of features by group/channel for \textbf{A.} the unlabelled model (\textit{Palette}), \textbf{B.} perturbation as a label (AdaGN and CG) and \textbf{C.} target as a label (AdaGN and CG). These features are extracted from the active subset. The number of features for each feature group is also presented (total 635 features). The mean correlations for all the selected features to the ground truth features are \textbf{A.} 0.386, \textbf{B.} 0.504, \textbf{C.} 0.355.}
  \label{correls}
\end{figure*}

We present CellProfiler feature correlation matrices between the features extracted from model predicted images and the features from the ground truth Cell Painting in Fig. \ref{correls}, which are presented as heatmaps. We compare three models: the unlabelled model, and models with perturbation and target as the label for the active subset. The labels were used in both training through AdaGN and sampling with classifier guidance. 

CellProfiler features are categorised as different feature groups (texture, radial distribution, intensity, granularity. colocalization, neighbours and area/shape) across the cells, cytoplasm and nuclei \cite{carpenter2006}. The correlation heatmaps present features after standard feature selection (total of 635 features), which includes dropping highly correlated features and zero-value features. Some feature groups in certain channels have no remaining features after feature selection (nan).

\subsection{Brightfield vs Cell Painting}


In this study it is notable how well the brightfield images perform in the transfer learning tasks for target matching. This was a surprising result given the limited studies in the literature which employ brightfield for image-based profiling. Although this behaviour could be unique to our dataset, these result pose a challenge to the utility of (fluorescent) label-free Cell Painting methods, which can be computationally intensive and may not necessarily outperform the brightfield modality in its own right. 

There are a number of advantages to imaging and profiling without fluorescent staining. Brightfield imaging is cheaper, requires minimal preparation, and does not damage the cells with photo- or cyto-toxic effects. In fluorescent staining, certain combinations of dyes are restricted due to the particular wavelength the dye can be imaged at (spectral overlap). These technical limitations can hinder the ability of the scientist to capture morphological information from the unstainable subcellular compartments.  Because of this, there is interest in using cheaper, quicker, less damaging alternatives such as brightfield to perform high-throughput screening and image-based profiling.

We investigate some of the quantitative and qualitative differences between brightfield, Cell Painting and our predicted Cell Painting (from brightfield) in this section. 

\subsubsection*{Overlap in matching target predictions}

\begin{figure}
  \centering
  \includegraphics[width=0.45\textwidth]{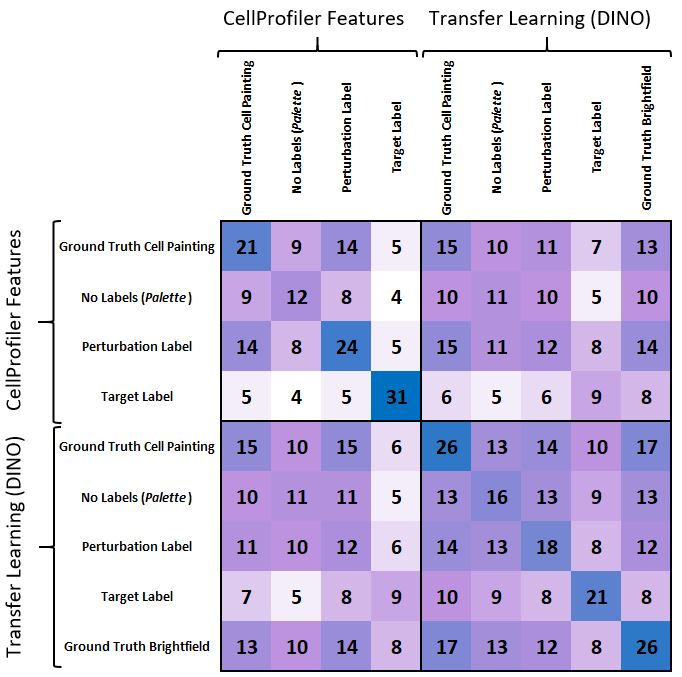}
  \caption[sc3]{Matrix of the total number of shared matching targets (NN  top 5) predicted between each of the models/ground truth modalities, in both CellProfiler Feature space and transfer learning (DINO) feature space. From the active subset, with labels included through AdaGN and CG.}
  \label{correls13}
\end{figure}

\begin{figure}
  \centering
  \includegraphics[width=0.45\textwidth]{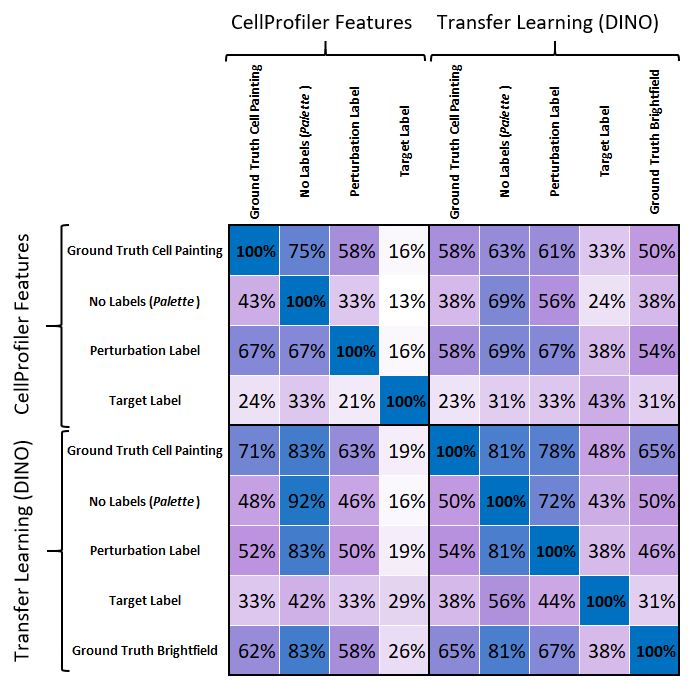}
  \caption[sc2]{Matrix of the values from Fig. \ref{correls13} expressed as a percentage of the total number of the value of the diagonal in the same column. i.e. the values in the first column are the \% of targets predicted by each model as a percentage of the targets predicted in the CellProfiler feature space extracted from the ground truth Cell Painting images. }
  \label{correls12}
\end{figure}

Firstly, we compare the overlap of the specific matching targets in each of the feature spaces of the different sets of images produced by the models. We also compare the ground truth Cell Painting and brightfield against the model predictions. For two or three models, this would be visualised with a Venn diagram. As we have more than three models to compare, we present the overlapping matching target predictions between models as two matrices in Figs. \ref{correls13} and \ref{correls12}. We used the active subset study for this analysis.

We find that there is a good overlap between the matching targets found by Cell Painting (both CellProfiler and DINO) and brightfield. When using the perturbation as the guiding label, comparable performance and reproducibility of matching targets is seen, however most of the other models produce worse results. Transfer learning with DINO generally produced over 50\% overlap in matching targets to the CellProfiler features, while also being capable of finding matching targets not in CellProfiler space. Perhaps a combined feature space (CellProfiler and transfer learning) could outperform the best individual models.

Although using the target as the label produced the largest number of matching targets (31), it shared very few matching targets to the other models including the ground truth. Even though it may be an advantage that this model can find different matching targets to the ground truth features, it should not come at the cost of failing to predict the simple-to-predict matching target pairs. Hence, we propose that using this label has not produced reproducible or correct features, rather the model has ``brute-forced'' similarities between images with the same labels, most likely by adding noise. We discuss this further in the Section \ref{bgnoise}.

\subsubsection*{Self-attention maps}

\begin{figure*}
  \centering
  \includegraphics[width=1\textwidth]{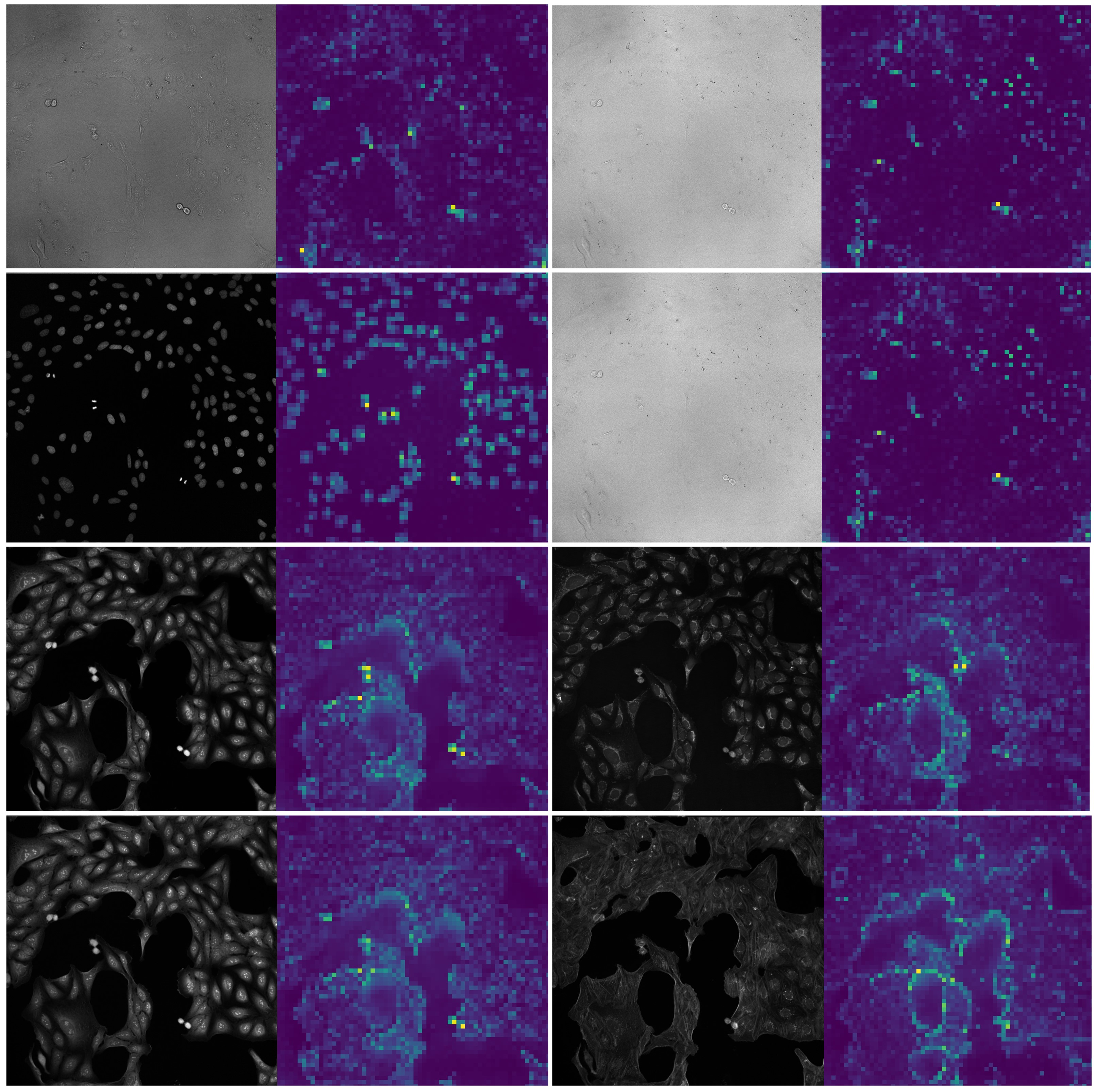}
  \caption[Example of self-attention heads for ground truth images]{An example of paired self-attention maps for ground truth images with transfer learning (DINO) weights. Left (top to bottom): Brightfield 1, DNA, RNA, ER. Right (top to bottom): Brightfield 2, Brightfield 3, Mito, AGP.}
  \label{lch62234}
\end{figure*}

In Fig. \ref{lch62234} we present self-attention maps for each of the ground truth channels using pretrained DINO weights \cite{caron2021}. Self attention maps provide a visualisation of which $8 \times 8$ patches the vision transformer network places most emphasis on when calculating a feature representation of the image. While the Cell Painting channels' self-attention maps are slightly sharper in their segmentation properties, the brightfield channels show fairly reliable segmentation of cellular structure. It may be the case that modern computer vision architectures such as self-supervised, attention-based transformer networks have unlocked the brightfield as a valid modality for image-based profiling. This has not been possible previously due to the lower resolution and higher noise of brightfield images. Our findings, alongside other recent studies \cite{gupta2022} provide motivation for further work investigating image-based profiling with brightfield images.

\begin{figure*}
  \centering
  \includegraphics[width=1\textwidth]{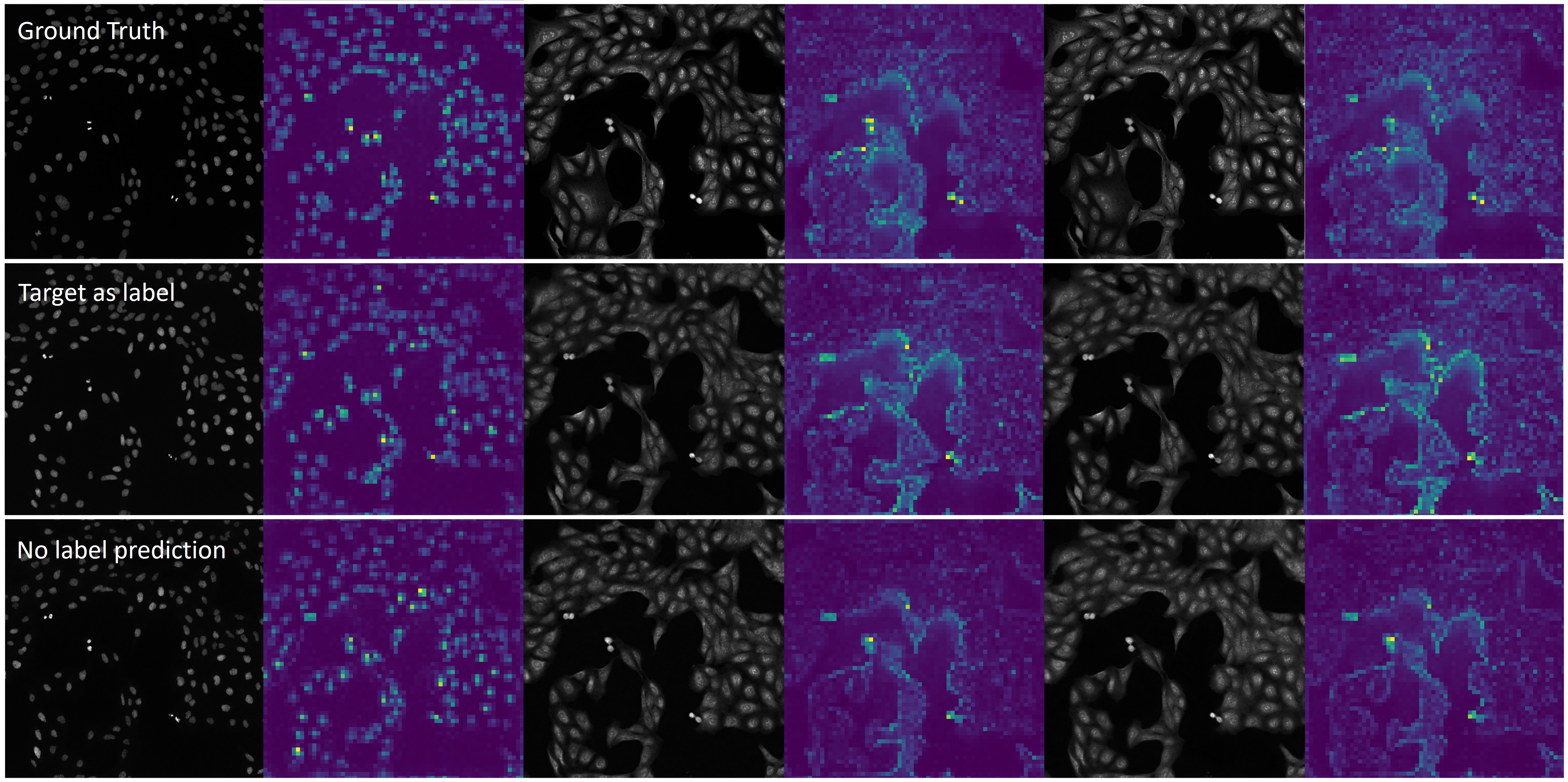}
  \caption[Examples of images and paired self attention maps for the ground truth vs diffusion model]{Examples of images and paired self attention maps for the ground truth vs diffusion model with label and without. Left to right: DNA, RNA, ER channels. The remaining two channels are displayed in Fig. \ref{lch64343}}
  \label{lch633214}
\end{figure*}

\begin{figure*}
  \centering
  \includegraphics[width=0.7\textwidth]{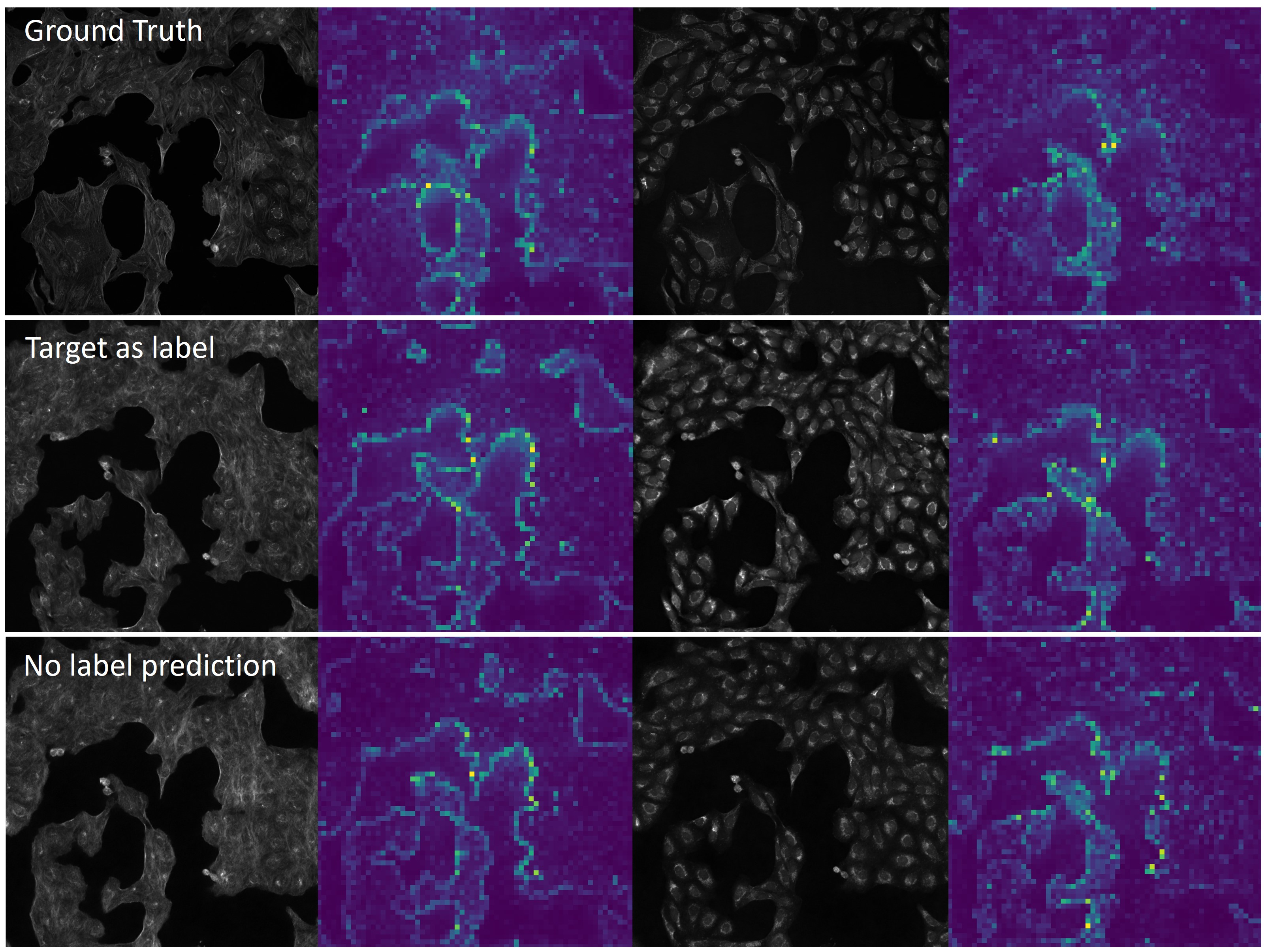}
  \caption[Examples of images and paired self attention maps for the ground truth vs diffusion model]{A continuation of Fig. \ref{lch633214}. Left to right: AGP, Mito.}
  \label{lch64343}
\end{figure*}

Additionally, in Figs. \ref{lch633214} and \ref{lch64343} we compare the self-attention maps of the predicted Cell Painting channels to the ground truth channels. While self-attention maps with $8 \times 8$ patches do not reveal fine-scale structure, we can see that the larger scale structural properties of the channels are replicated relatively well in the predicted Cell Painting channels. 

\subsection{Background noise}\label{bgnoise}

\begin{figure*}
  \centering
  \includegraphics[width=1\textwidth]{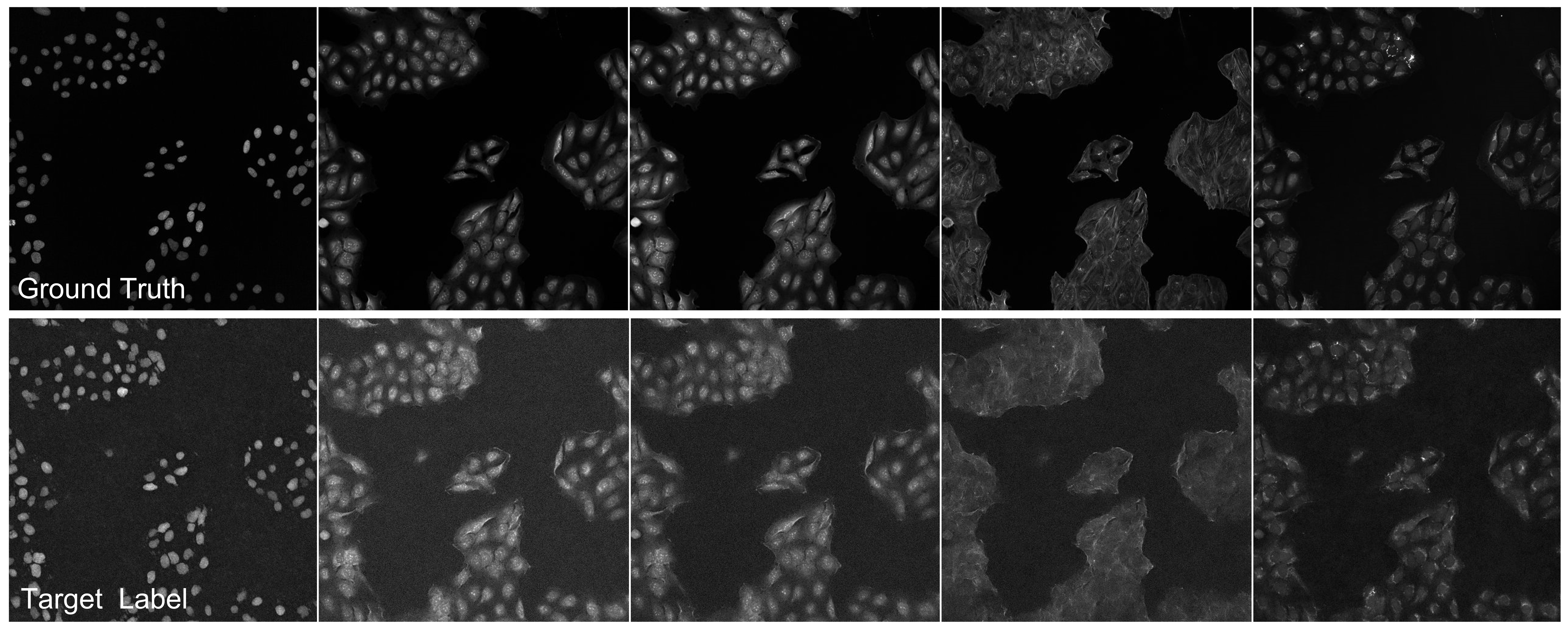}
  \caption[Example of all channels (and input) for both modelss]{When target is used as a label in training and inference, sometimes background noise is added to the images when the model is unable to output a structural improvement. Columns left to right: DNA, RNA, ER, Mito, AGP.}
  \label{lch61657}
\end{figure*}

We noticed that some of the predicted images were noisy across the whole image, which was particularly visible in the background (Fig. \ref{lch61657}). The model would not add noise to all the predicted images in the test set, just a small number with certain class labels. This could be seen as a form of overfitting, where the model has learned to output irrelevant noise patterns which make images of the same class more similar. This is reflected in the metrics in Tables \ref{tab:tab1} and \ref{tab:tab2}, where target matching improves despite reconstruction quality dropping in all metrics. 

This was most common with target as the class label. One way to think about this is that if we were to construct a classifier to predict the target from the input images, this would be a very difficult task (in fact, this simple problem motivates much of the field of image-based profiling in drug discovery) compared to using the perturbation. Hence we emphasise the importance of a sensible choice of class labels. We should not introduce labels which are too ambitious for the network, and which may prevent learning a faithful reconstruction. Instead, we propose that the utility of class-guided image-to-image diffusion is through using simple labels to guide learning important structural fundamentals.

\subsection{Experimental batch effects}

The experimental batch effect is always a consideration in image-based profiling, and many studies have focused on tackling it \cite{Lin2022, Wang2021, ando2017}. We provide a few remarks relating to our study and the batch effect in this section.

\begin{figure*}
  \centering
  \includegraphics[width=1\textwidth]{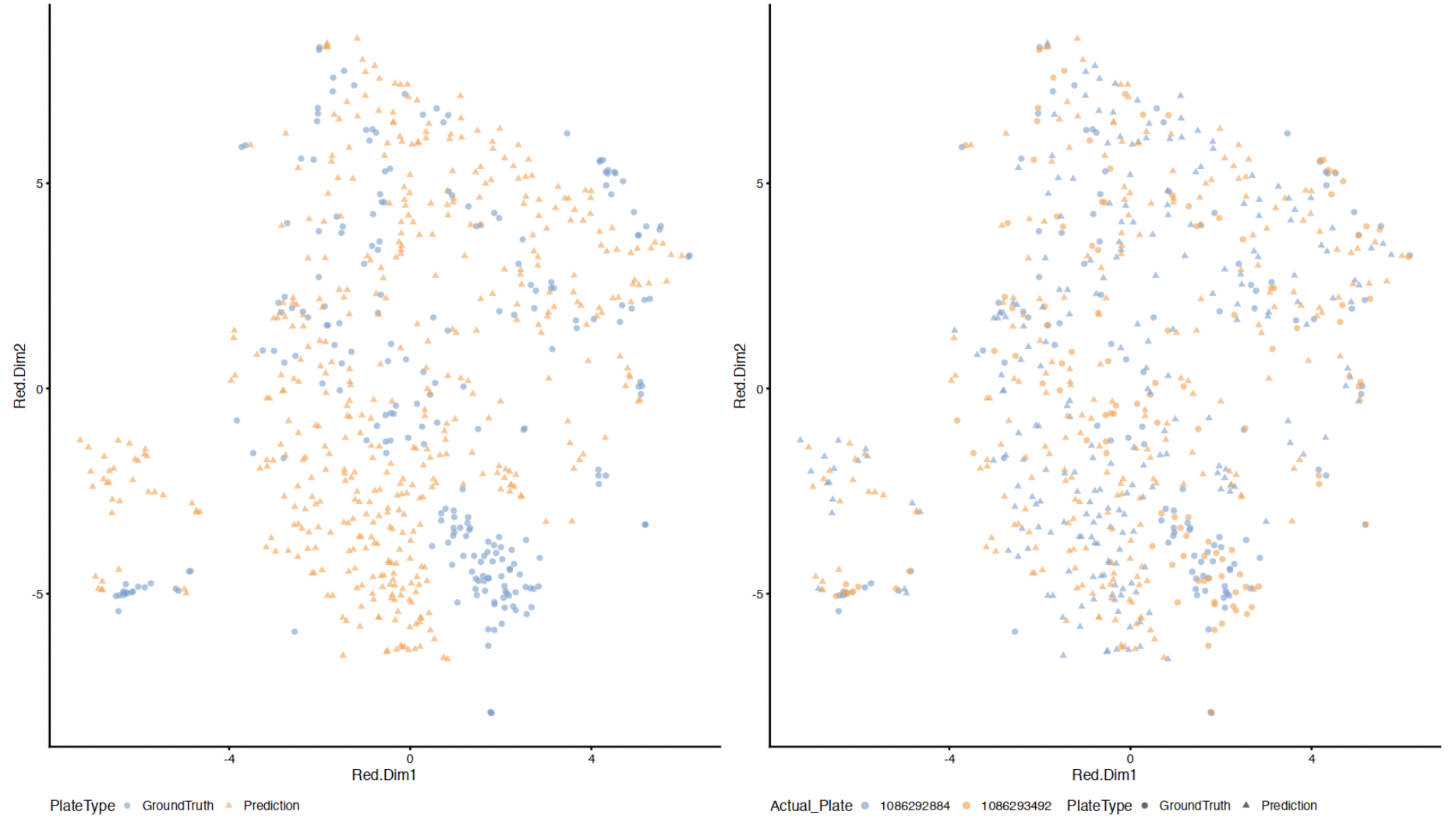}
  \caption[t-SNE plots of both plat3es]{t-SNE plots of both plates for the unlabelled model. There is some separation between the real and predicted spaces, but also signifcant overlap. There is no batch effect between the plates in either the ground truth or predicted plates.}
  \label{lch66839}
\end{figure*}

If there is a batch effect in the ground truth, a faithful reconstruction would preserve it. Since we are producing potentially entire plates of images, all batch correction/normalisation methods which can be applied to real Cell Painting can also be applied to the predicted images (for example TVN \cite{ando2017} which uses variation in DMSO controls to correct for experimental batch variation). For this reason, we intentionally tested each replicate of our models on a single test batch. However, we present a model trained on 8 plates and simultaneously tested on 2 unseen plates to study the batch effect. We present a 2-D t-SNE plot of this investigation in Fig. \ref{lch66839}. There was no notable batch effect between our ground truth plates, and we saw this replicated in our models (with and without labels). Promisingly, the CellProfiler feature spaces extracted from model predicted images overlapped well with the ground truth feature space. This is an improvement upon prior studies, which induced a phantom ``batch effect'' between the predicted and real feature spaces \cite{CrossZamirskiGAN}. This is important to asses, as correlation (Fig. \ref{correls}) does not account for feature space overlap.

\end{document}